\documentclass{article}
\usepackage{spconf,amsmath,graphicx}
\newcommand{\norm}[1]{\left\lVert#1\right\rVert}

\title{Bone Structures Extraction and Enhancement in Chest Radiographs \\
via CNN trained on Synthetic data }
\name{Ophir Gozes, Hayit Greenspan }
\address{Department of Biomedical Engineering, \\ Faculty of Engineering, Tel Aviv University, 
	  Israel}
\begin{document}
%
\maketitle
\begin{abstract}

In this paper, we present a deep learning-based image processing technique for extraction of bone structures in chest radiographs using a U-Net FCNN.
The U-Net  was trained to accomplish the task in a fully supervised setting.
To create the training image pairs, we employed simulated X-Ray or Digitally Reconstructed Radiographs (DRR), derived from 664 CT scans belonging to the LIDC-IDRI dataset. 
Using HU based segmentation of bone structures in the CT domain, a synthetic 2D "Bone x-ray" DRR is  produced and used for training the network.
For the reconstruction loss, we utilize two loss functions- L1 Loss and perceptual loss. 
Once the bone structures are extracted, the original image can be enhanced by fusing the original input x-ray and the synthesized ``Bone X-ray". We show that our enhancement technique is applicable to real x-ray data, and display our results on the NIH Chest X-Ray-14 dataset.

\end{abstract}
\section{Introduction}
\label{sec:intro}

Chest X-ray (CXR) 
is the most frequently performed diagnostic X-ray examination. It produces images of the heart, lungs, airways, blood vessels and the bones of the spine and chest.
Bones are the most dense structures visible on a normal Chest X-ray and include the ribs, clavicles,the scapulae, part of the spine and the upper arms.
The ribs are essential structures of the osseous thorax and provide information that aids in the interpretation of radiologic images. Techniques for making precise identification of the ribs are useful in detection of rib lesions and localization of lung lesions \cite{C1}. The large number of overlapping anatomical structures appearing in a CXR can cause the radiologist to overlook bone pathologies.

\par

One of the motivations for extraction of the rib bones is to allow their suppression \cite{C4}. This allows obtaining soft-tissue-like images for better soft lung structure analysis. By using Dual Energy acquisition it is possible to achieve comparable results, but since it requires a dedicated scanner it is not commonly performed.
In 2006 Suzuki et al.\cite{C3} introduced a method for suppression of ribs in chest radiographs by means of Massive Training Artificial Neural Networks (MTANN). Their work relied on Dual energy X-ray in the creation of training images. They named the resulting images “bone-image-like” and by subtracting them they were able to produce “soft-tissue-image-like” images where ribs and clavicles were substantially suppressed. The bone images created in the process were limited to clavicles and ribs, and were relatively noisy compared to the dual-energy bone image.

Other recently published works have used Digitally Reconstructed Radiographs (DRR) for training CNN models. Esteban et al.\cite{C10}  used DRR for training X-ray to CT registration while Albarqouni et al. \cite{C6} used DRR image training for decomposing CXR into several anatomical planes.
In \cite{C4} authors used rib-bone atlases to automatically extract the patient rib-bone on conventional frontal CXRs. Their  system chooses the most similar models in the atlas set and then registers them to the patient’s X-ray. In recent work \cite{C5} a Wavelet-CNN model for bone suppression was introduced and used to predict a series of wavelet sub-band images of bone image. They used dual energy subtraction (DES) CXRs for training.

In earlier work \cite{C2}, we presented a method for enhancing the contrast of soft lung structures in chest radiographs using U-Nets. The current work provides an extension by addressing specifically the bones. Most notable difference to previous works is the use of  perceptual loss \cite{C7} for improved reconstruction of the bone image. As an output of the work we focus  on enhancement of bone structures in chest X-ray. Our training approach is based on CT data and is focused on extraction of bones and their enhancement in combination with the original radiograph. The enhanced result maintains a close appearance to a regular x-ray, which may be attractive to radiologists.

	
	
	
An overview of our approach is presented in 
Section 2. Experimental results are shown in Section 3. In Section 4, a discussion of the results is presented followed by a conclusion of the work.

\section{Methods}
\label{sec:methods}

\begin{figure}[htb]
			\centerline{\includegraphics[width=8.0cm]{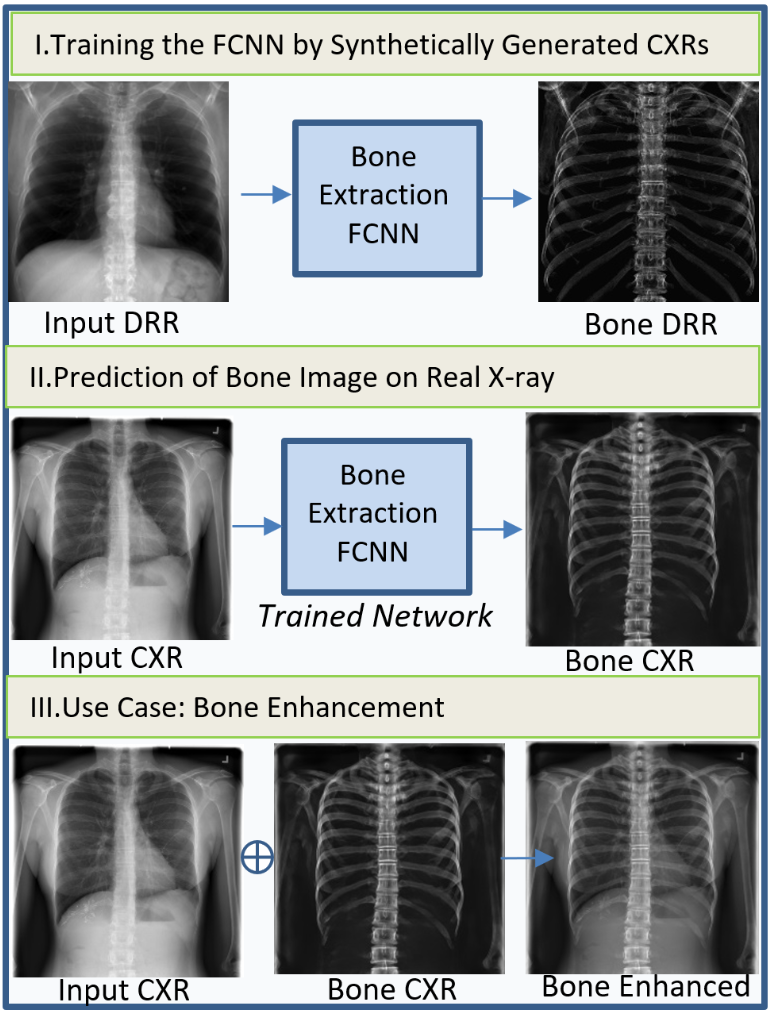}}
	\caption{Description of our Approach}
	 \label{method_scheme}
\end{figure}

Given a chest X-ray as input, we introduce a system that allows extraction of bone structures as well as synthesis of an enhanced radiograph. An overview of the approach is presented in Fig.\ref{method_scheme}.
The proposed methodology is based on  neural networks trained on synthetic data (Fig.\ref{method_scheme}.I)
To produce training images we use DRRs that were  generated from a subset of LIDC-IDRI dataset \cite{C11}.
Following training, we use the trained network on a real CXR to perform prediction of bone image (Fig.\ref{method_scheme}.II)
Finally, we present a use case in which the bone image is used for bone enhancment on a real CXR (Fig.\ref{method_scheme}.III)

%

\par

\subsection{Synthetic X-ray Generation (DRRs) }\label{drr_section}
We  review the physical process which governs chest X-ray generation and  our method for simulating synthetic X-ray, or digitally reconstructed radiographs (DRRs). The method is based on our recently published work \cite{C2}:
As X-rays propagate through matter, their energy decreases.
This attenuation in the energy of the incident X-ray depends on the distance traveled and on the attenuation coefficient.
This relationship is expressed by Beer Lambert's law, where $I_{0}$ is the incident beam, I is the Intensity after traveling a distance x and A is the attenuation coefficient: 
\begin{equation}I = I_{0}\exp^{Ax}
\end{equation}
\par In order to simulate the X-ray generation process, calculation of the attenuation coefficient is required for each voxel in the CT volume. In a CT volume, each voxel is represented by its Hounsfield unit (HU) value, which is a linear transformation of the original linear attenuation coefficient. Therefore the information regarding the linear attenuation is maintained.
We assume for simplicity a parallel projection model and compute the average attenuation coefficient along the y axis ranging from [1,N] (where N is the pixel length of the posterior anterior view).
Denoting the CT volume by  G(x,y,z), the 2D average attenuation map can be computed using Eq. 2:
\begin{equation}
\mu_{av}(x,z) = \sum_{y=1}^{N}  \dfrac{\mu_{water}(G(x,y,z)+1000)}{(N\cdot 1000)}
\end{equation}
Utilizing Beer Lambert's law (Eq 1) the DRR is generated as shown in Eq. 3:
\begin{equation}
I_{DRR}(x,z) = \exp^{\beta \cdot\mu_{av}(x,z)} 
\label{DRR_create_eq}
\end{equation}
The attenuation coefficient of water $\mu_{water}$ was taken as 0.2 $  CM^{-1}$ while $\beta$ was selected as 0.02 such that the simulated X-ray matched the appearance of real X-ray images.

\subsection{Creation of 2D image pairs  for training: }
\begin{figure}[htb]

	\begin{minipage}[b]{.48\linewidth}
		\centering
		\centerline{\includegraphics[width=4.0cm]{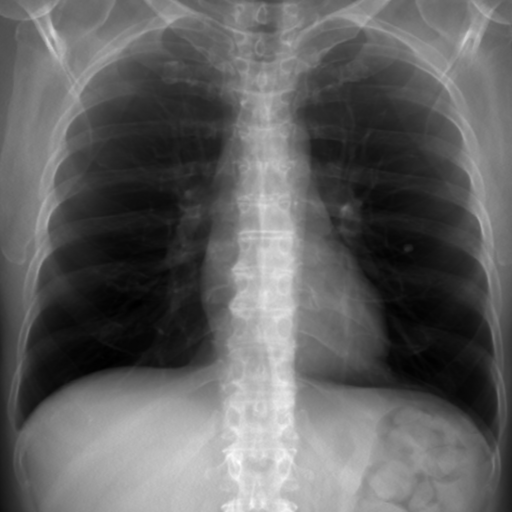}}
		\centerline{(a) Source - DRR  Image}\medskip
	\end{minipage}
	\hfill
	\begin{minipage}[b]{0.48\linewidth}
		\centering
		\centerline{\includegraphics[width=4.0cm]{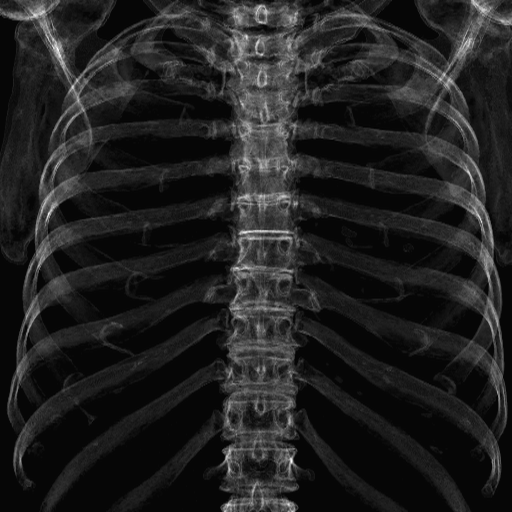}}
		\centerline{(b) Target - ``bone X-ray" }\medskip
	\end{minipage}
	\caption{synthetic image pair used for training}
	\label{fig:drr_bone1}
\end{figure}

Our goal is to extract the bone structures of a given chest X-ray. For this we employ a U-Net and train it using pairs of synthetic X-ray (DRR) images. 
The source image is a synthetic X-ray generated as described in section \ref{drr_section}.
For generating the target image, we first define bone voxels as the set of voxels with HU values in  [300,700]. We set all non bone voxels to a value of -1024 HU (Air). We then generate the synthetic X-ray (section \ref{drr_section}) corresponding to the Bone voxels.
We term the resulting target image:  ``bone X-ray" since all soft tissues are removed from this image.
An example ``bone X-ray'' is presented in Fig.\ref{fig:drr_bone1}. It is noticeable that only bone structures appear, excluding overlapping soft tissue anatomical structures such as the heart, liver, stomach and fat.

\subsection{Bone Extraction U-Net }
\begin{figure}[!htb]

	\begin{minipage}[b]{1\linewidth}
		\centering
		\centerline{\includegraphics[width=8cm]{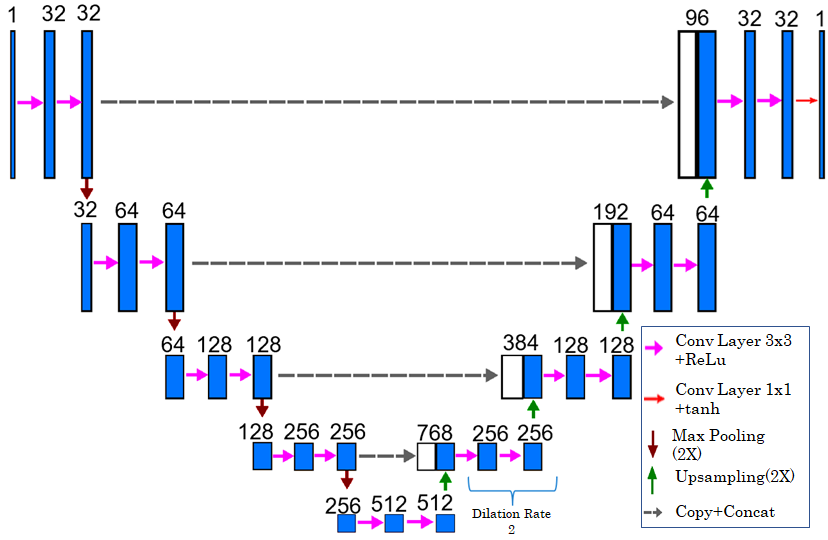}}
	\end{minipage}
	\caption{U-Net Architecture for Bone Extraction}
	\label{fig:unet}
\end{figure}

The network used in this work are based on the U-net FCNN architecture \cite{C8}. 
We specify here the modifications which we made to the original architecture:
The inputs and output size is $512\times512$ with 32 filters in the first convolution layer
Dilated convolutions (dilation rate = 2) were added in areas of the network in order to enlarge the receptive field.
We use ReLu activation functions throughout the net while at the network output we use the Tanh activation.
A diagram describing the architecture is presented in Fig. \ref{fig:unet}

\begin{figure}[htb]

			\centerline{\includegraphics[width=8.0cm]{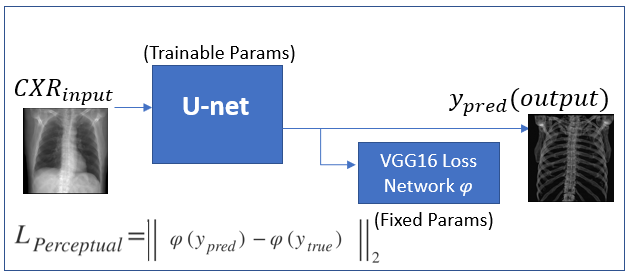}}
	\caption{Description of the training Scheme}
	 \label{training_scheme}
\end{figure}

\subsubsection{Loss Functions  }
We examined two loss functions, the first is the commonly used L1 Loss, the second is the perceptual loss. The L1 loss aims to minimize pixel level difference, while Perceptual loss encourages the reconstruction to have a similar feature representation.  

The L1 loss  is a point loss which minimizes the gray value absolute difference, as in Eq.\ref{l1_loss}. 
\begin{equation}
L_{L1}=\norm{(y_{pred} - (y_{gt})}_{1}
\label{l1_loss}
\end{equation}

The perceptual loss measures high-level semantic differences between images. It makes use of a loss network, pre-trained
for image classification. The loss network $\varphi$ used in this work is a VGG16 CNN, pretrained on ImageNet.
For perceptual loss, we compute the feature reconstruction loss, as in  Eq.\ref{perceptual_loss}. We denote the activation output of layer $'block2\_conv2'$ (the fourth layer) of  VGG16 network as $\varphi(x)$.
In order to fit the input requirements of the VGG16 network which was trained on 224$\times$224$\times$3 color images, we 
duplicate the predictions of the network to create an RGB output denoted as  $yc_{pred}$ ,$yc_{gt}$ resulting in image of dimensions 512$\times$512$\times$3.
The training scheme is displayed in Fig.\ref{training_scheme}.

\begin{equation}
L_{Perceptual}=\norm{(\varphi(yc_{pred}) -\varphi( yc_{gt})}_{2}
\label{perceptual_loss}
\end{equation}

\subsubsection{Training Details }

For training, we used zero mean normalized batches of size 8 and ADAM optimizer. The optimal initial learning rate was found to be  1E-3. Validation loss converged after 100 epochs.

We utilized random data augmentation for both the source and the target using horizontal flipping, additive Gaussian noise, bias($\pm20\%$), scaling($\pm30\%$) and image sharpening(alpha=0.5).

\subsection{Use Case: A Scheme for bone Structures Enhancement in Real CXR}
Once extracted, the bone structures can be added to the original DRR image, allowing for a selective enhancement of bone structures.
For the enhancement of a real radiograph, we use the bone structures extraction U-Net to extract bone structures from the input image (i.e prediction of a ``bone X-ray").
We then fuse the two images using a weighted summation.

\section{Experiments and Results}

{\bf{Bone Structures Extraction Metrics on Synthetic Data}}

A subset of 644 CT scans belonging to LIDC dataset was used.
For each CT case we generate a DRR and a ``bone X-ray" pair.
We split the dataset to training, validation and test subsets containing 386, 129, 129  pairs, respectively.
We evaluate our results on the test subset and report  RMSE, PSNR, SSIM(Structural similarity) and MSSIM(multiscale SSIM).
Results are given in Table.\ref{metrics}. We see that in the  synthetic data scenario, the best metrics were achieved using perceptual loss training: MSSIM: 0.84, SSIM: 0.7,PSNR: 22.6dB and RMSE :19.25. 

\renewcommand{\tabcolsep}{1pt}
\begin{table}[!h]
	%
	\centering
	\caption{Bone Extraction Test Metrics on Synthetic X-ray (mean,std)}\label{tab2}
	\begin{tabular}{c|c|c|c|c}
		\hline
		Loss&RMSE& PSNR[dB] &SSIM &MSSIM\\
		\hline
		{L1} &    \text{20.5(4.1)} &   \text{22.0(1.6)} &
		\text{0.70(.07))} &   \text{0.83(0.04)} \\
		\hline
		{Perceptual} &    \textbf{19.25(4)} &   \textbf{22.6(1.7)} &
		\textbf{0.70(.08)} &   \textbf{0.84(.04)} \\
		\hline
	\end{tabular}
	\label{metrics}
\end{table}

{\bf{Applicability to Real X-ray}}\\
In order to explore the applicability of our algorithm to real X-ray, we apply the bone extraction U-Net on images from NIH ChestX-ray14 dataset \cite{C9}. The results were obtained by using perceptual loss during training.

\begin{figure}[!htb]

	\begin{minipage}[b]{.48\linewidth}
		\centering
		\centerline{\includegraphics[width=4.0cm]{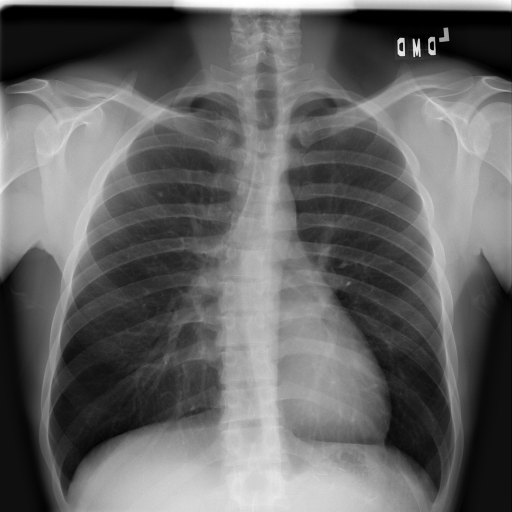}}
		\centerline{(a) Real CXR}\medskip
	\end{minipage}
	\hfill
	\begin{minipage}[b]{0.48\linewidth}
		\centering
		\centerline{\includegraphics[width=4.0cm]{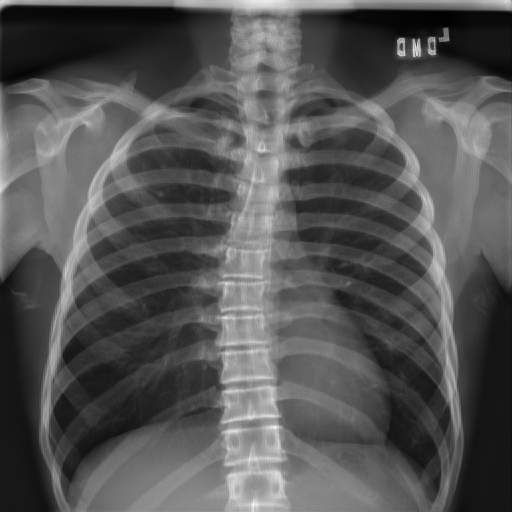}}
		\centerline{(b) Bone Enhanced CXR}\medskip
	\end{minipage}
	\begin{minipage}[b]{.48\linewidth}
	\centering
	\centerline{\includegraphics[width=4.0cm]{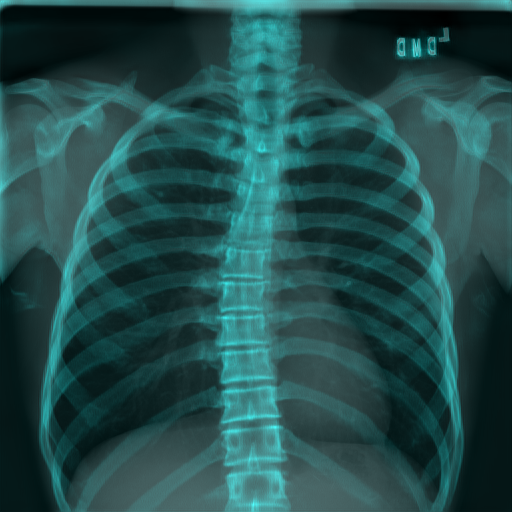}}
	\centerline{(c) Color View}\medskip
\end{minipage}
\hfill
\begin{minipage}[b]{0.48\linewidth}
	\centering
	\centerline{\includegraphics[width=4.0cm]{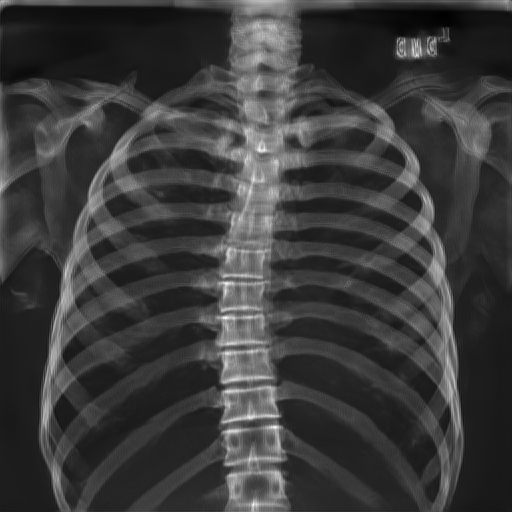}}
	\centerline{(d) Extracted Bones}\medskip
\end{minipage}
		
		\caption{Real X-ray Results - NIH X-ray14 }
		\label{fig:xray_res1}
		
	\end{figure}

In Figure \ref{fig:xray_res1} we show the results of the enhancement on a Real X-ray. It is noticeable that the heart and other soft structures are removed from the  extracted bone layer.

\begin{figure}[!htb]


	\begin{minipage}[b]{.48\linewidth}
	\centering
		\centerline{\includegraphics[trim=70 70 40 200,clip, width=4.0cm]{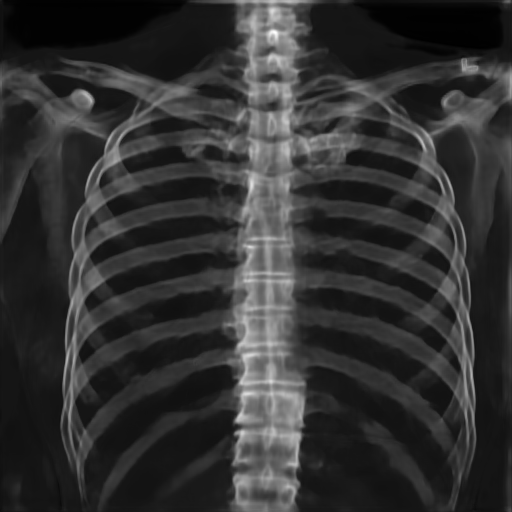}}
	\centerline{(a)L1 Loss(zoomed view)}\medskip
\end{minipage}
\hfill
\begin{minipage}[b]{0.5\linewidth}
	\centering
	\centerline{\includegraphics[trim=70 70 40 200,clip, width=4.0cm]{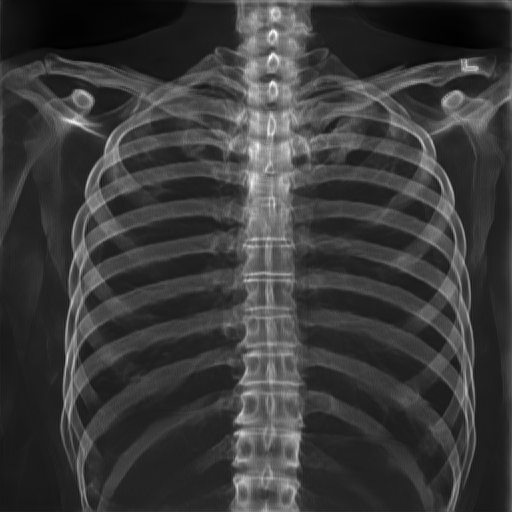}}
	\centerline{(b) Perceptual Loss(zoomed view)}\medskip
\end{minipage}
	\caption{L1 Loss Vs Perceptual Loss - NIH X-ray14 }
	\label{fig:xray_l1_vs_perceptual}
	
\end{figure}

{\bf{Real X-ray: Perceptual Loss vs L1 Loss - Qualitative }}\\
In Fig.\ref{fig:xray_l1_vs_perceptual}, two bone images were created by two U-Net, one trained using L1 and a second trained by Perceptual Loss. Using Perceptual Loss we achieved a higher level of detail in the vertebrae, and the image sharpness is improved.

\section{Discussion and Conclusion}
In this work, we  presented a method for enhancement and extraction of bone structures in chest X-ray.
We used synthetic data for training and showed that the resulting trained networks are robust even for Real CXRs.
The use of perceptual loss showed improved metrics on the synthetic dataset while for real CXR, it improved sharpness and detail of the bone images.

\bibliographystyle{IEEEbib}

\bibliography{bib_shorten}

\end{document}